\documentclass{article}

\usepackage{PRIMEarxiv}

\usepackage[utf8]{inputenc} 
\usepackage[T1]{fontenc}    
\usepackage{hyperref}       
\usepackage{url}            
\usepackage{booktabs}       
\usepackage{amsfonts}       
\usepackage{nicefrac}       
\usepackage{microtype}      
\usepackage{lipsum}
\usepackage{fancyhdr}       
\usepackage{graphicx}       
\graphicspath{{media/}}     
\usepackage{enumitem}
\usepackage{multirow}
\usepackage{graphicx}
\usepackage{caption}
\usepackage{amsmath}
\usepackage{colortbl}
\usepackage{mathtools}
\usepackage{tikz}

\definecolor{blue-violet}{rgb}{0.54, 0.17, 0.89}
\definecolor{bluegray}{rgb}{0.4, 0.6, 0.8}
\definecolor{bleudefrance}{rgb}{0.19, 0.55, 0.91}
\definecolor{darkblue}{rgb}{0.0, 0.0, 0.55}
\definecolor{denim}{rgb}{0.08, 0.38, 0.74}
\definecolor{mediumblue}{rgb}{0.0, 0.0, 0.8}
\definecolor{naplesyellow}{rgb}{0.98, 0.85, 0.37}
\definecolor{carminered}{rgb}{1.0, 0.0, 0.22}
\definecolor{crimsonglory}{rgb}{0.75, 0.0, 0.2}
\definecolor{lightgreen}{rgb}{0.56, 0.93, 0.56}
\definecolor{lightapricot}{rgb}{0.99, 0.84, 0.69}
\definecolor{pastelred}{rgb}{1.0, 0.41, 0.38}
\definecolor{periwinkle}{rgb}{0.8, 0.8, 1.0}
\definecolor{xgreen}{rgb}{0.88, 0.95, 0.83}
\definecolor{uared}{rgb}{0.85, 0.0, 0.3}
\definecolor{debianred}{rgb}{0.84, 0.04, 0.33}
\definecolor{red(ncs)}{rgb}{0.77, 0.01, 0.2}
\definecolor{blush}{rgb}{0.87, 0.36, 0.51}
\definecolor{non-photoblue}{rgb}{0.64, 0.87, 0.93}
\definecolor{lightcornflowerblue}{rgb}{0.6, 0.81, 0.93}
\definecolor{cadmiumgreen}{rgb}{0.0, 0.42, 0.24}
\definecolor{dollarbill}{rgb}{0.52, 0.73, 0.4}
\definecolor{darkgreen}{rgb}{0.0, 0.2, 0.13}
\newcommand\medvidcl{\texttt{MedVidCL}}
\newcommand\medvidqa{\texttt{MedVidQA}}
\newcommand\vslbase{\textsc{VSL-Base}}
\newcommand\vslqgh{\textsc{VSL-Qgh}}
\title{A Dataset for Medical Instructional Video Classification and Question Answering}

\author{
  Deepak Gupta, Kush Attal, and Dina Demner-Fushman \\
  Lister Hill National Center for Biomedical Communications \\
  National Library of Medicine, National Institutes of Health \\
  Bethesda, MD, USA\\
  \texttt{\{firstname.lastname\}@nih.gov} 
}

\begin{document}
\maketitle

\begin{abstract}
This paper introduces a new challenge and datasets to foster research toward designing systems that can understand medical videos and provide visual answers to natural language questions. We believe medical videos may provide the best possible answers to many first aid, medical emergency, and medical education questions. Toward this, we created the \medvidcl{} and \medvidqa{} datasets and introduce the tasks of Medical Video Classification (MVC) and Medical Visual Answer Localization (MVAL), two tasks that focus on cross-modal (medical language and medical video) understanding. The proposed tasks and datasets have the potential to support the development of sophisticated downstream applications that can benefit the public and medical practitioners.
Our datasets consist of $6,117$ annotated videos for the MVC task and $3,010$ annotated questions and answers timestamps from $899$ videos for the MVAL task. These datasets have been verified and corrected by medical informatics experts. We have also benchmarked\footnote{\url{https://github.com/deepaknlp/MedVidQACL}} each task with the created \medvidcl{}  and \medvidqa{} datasets and propose the multimodal learning methods that set competitive baselines for future research. 
\end{abstract}

\keywords{Multimodal Learning \and Video Localization \and Video Classification}

\section{Introduction}
One of the key goals of artificial intelligence (AI) is developing a multimodal system that facilitates communication with the visual world (i.e., images, videos) using a natural language query. In recent years, significant progress has been achieved due to the introduction of large-scale language-vision datasets and the development of efficient deep neural techniques that bridge the gap between language and visual understanding. Improvements have been made in numerous vision-and-language tasks, such as visual captioning \cite{li2020oscar,luo2020univl}, visual question answering \cite{zhang2021vinvl}, and natural language video localization \cite{anne2017localizing,hendricks2018localizing,Liu2018AMR}. In recent years there has been an increasing interest in video question-answering \cite{lei2018tvqa,lei2019tvqa+} tasks,  where given a video, the systems are expected to retrieve the answer to a natural language question about the content in the video. We argue that only predicting the natural language answer does not seem to reflect the real world, where people interact through natural language questions and expect to localize the moments from the videos to answer their questions. The majority of the existing work on video question answering (VQA) focuses on \textbf{(a)} open-domain applications by building the VQA datasets \cite{tapaswi2016movieqa,lei2019tvqa+,mun2017marioqa} consisting of movies, TV shows, and games, and \textbf{(b)} retrieval \cite{lei2018tvqa,lei2019tvqa+,tapaswi2016movieqa} of the natural language answers.  With increasing interest in AI to support clinical decision-making and improve patient engagement \cite{hhsAI}, there is a need to explore such challenges and develop efficient algorithms for medical language-video understanding.

\begin{figure}
    \centering
    \includegraphics[width=0.9\linewidth]{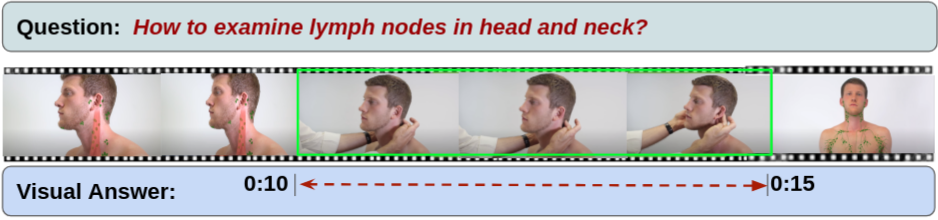}
    \caption{{A sample example of a health-related question and its visual answer (temporal segment) from the video.}}
    \label{fig:example}
\end{figure}

The recent surge in availability of online videos has changed the way of acquiring information and knowledge. Many people prefer instructional videos to teach or learn how to accomplish a particular task with a series of step-by-step procedures. 
Medical instructional videos are more suitable and beneficial to deliver the key information through visual and verbal communication at the same time in an effective and efficient manner. Consider the following medical question: ``\textit{How to place a tourniquet in case of fingertip avulsions?}'' The textual answer to this question will be hard to understand and act upon without visual aid. To provide visual aid, we first need to identify the relevant video that is medical and instructional in nature. Once we find a relevant video, it is often the case that the entire video can not be considered as the answer to the given question. Instead, we want to refer to a particular temporal segment, or a sequence of moments, from the video, where the answer is being shown or the explanation is being illustrated. The straightforward moment retrieval \textit{via} an action, object, or attribute keyword may not uniquely identify the relevant temporal segment, which consists of the visual answer to the question. A more natural way to refer to the appropriate temporal segment (\textit{c.f.} Fig \ref{fig:example}) is via natural language question and video segment description, which requires a fine-grained semantic understanding of the video segment, segment description, and question.

\begin{figure}[!h]
    \centering
    \includegraphics[height=16cm, width=0.9\textwidth]{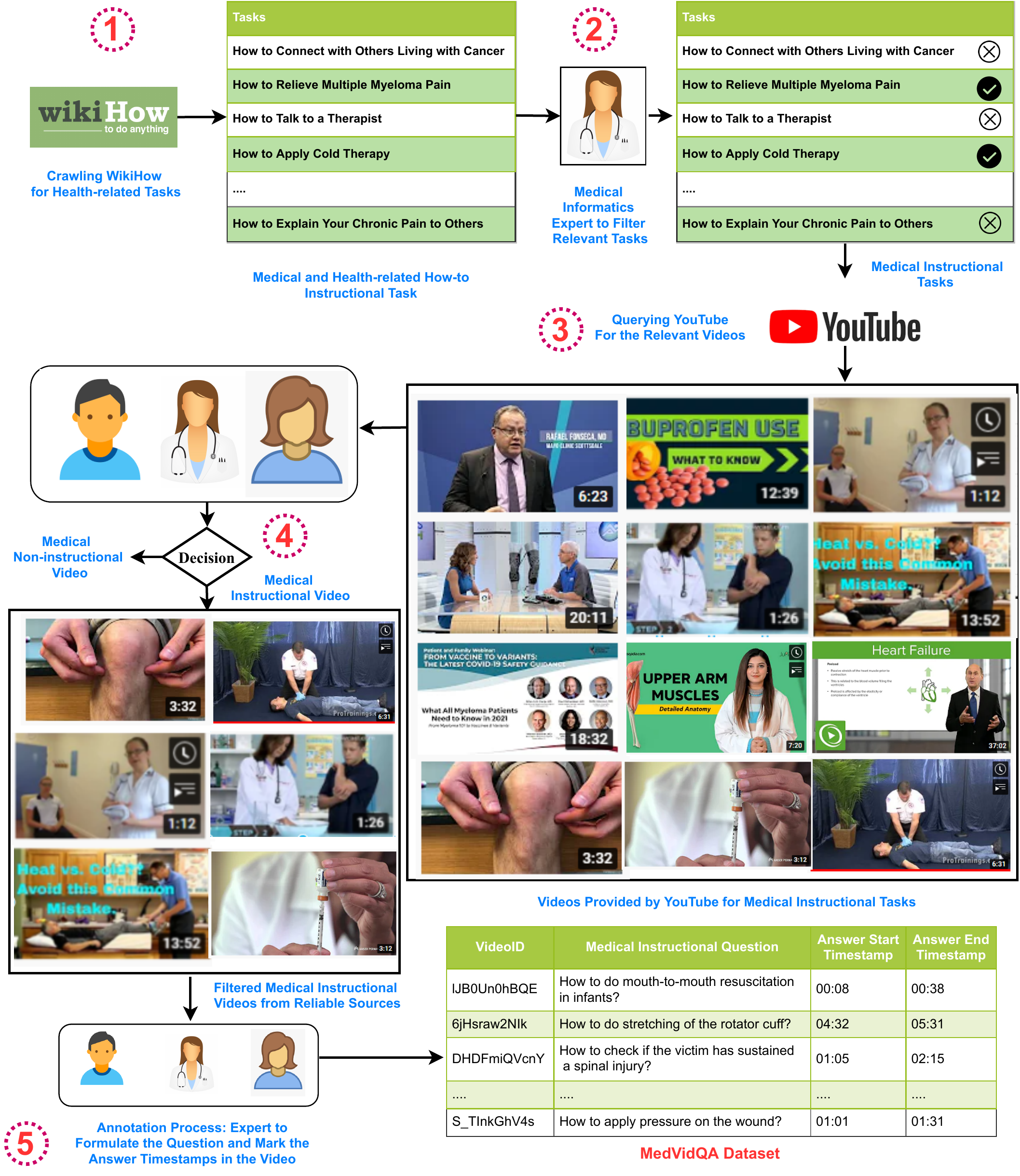}
    \caption{The schematic workflow of the \medvidqa{} dataset creation. Each step  \protect\begin{tikzpicture}[baseline=-1mm] \protect\draw[line width=0.35mm, uared, dotted] circle [radius=0.25] node {$\mathbf{N}$}; \protect\end{tikzpicture} is discussed in \medvidqa{} Data Creation.}
    \label{fig:medvidqa_dataset_overview}
\end{figure}

\begin{figure}[h]
    \centering
    \includegraphics[scale=0.6]{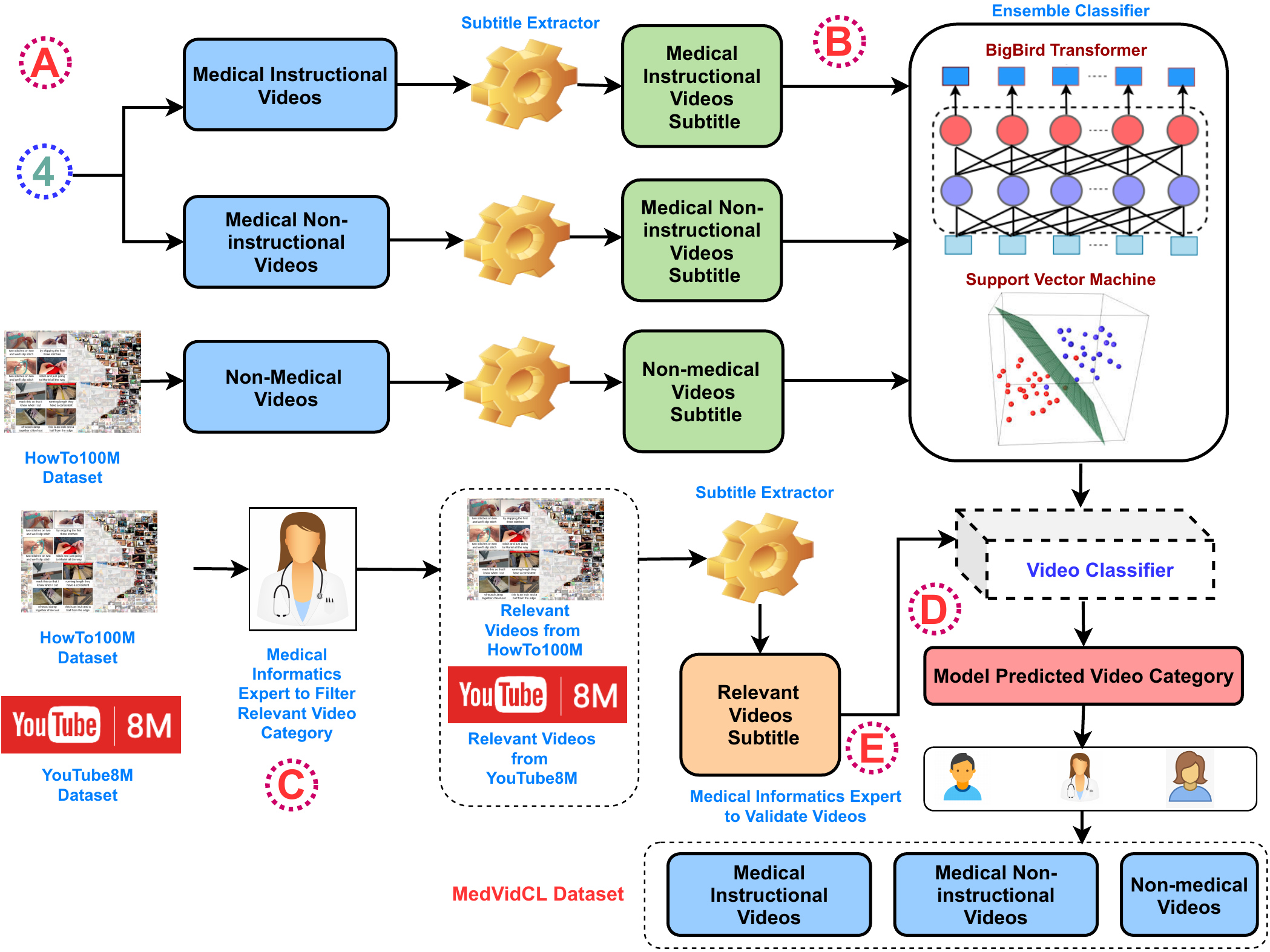}
    \caption{The schematic workflow of the \medvidcl{} dataset creation. It starts with the collection of medical instructional and non-instructional videos from step \protect\begin{tikzpicture}[baseline=-1mm] \protect\draw[line width=0.35mm, blue, dotted] circle [radius=0.25] node {$\textcolor{dollarbill}{\mathbf{4}}$}; \protect\end{tikzpicture} of the \medvidqa{} dataset creation. Each step  \protect\begin{tikzpicture}[baseline=-1mm] \protect\draw[line width=0.35mm, uared, dotted] circle [radius=0.25] node {$\mathbf{X}$}; \protect\end{tikzpicture} is discussed in \medvidcl{} Data Creation.}
    \label{fig:medvidcl_dataset_overview}
\end{figure}

We introduce the \textbf{Med}ical \textbf{Vid}eo \textbf{CL}assification (\medvidcl{}) and  \textbf{Med}ical \textbf{Vid}eo \textbf{Q}uestion \textbf{A}nswering (\medvidqa{}) datasets for medical instructional video classification and question answering. The \medvidcl{} dataset contains a collection of $6,617$ videos annotated into \textit{`medical instructional'}, \textit{`medical non-instructional'} and \textit{`non-medical'} classes. We adopted a two-step approach to construct the \medvidcl{} dataset. In the first step, we utilize the videos annotated by health informatics experts to train a machine learning model that predicts the given video to one of the three aforementioned classes. In the second step, we only use high-confidence videos and manually assess the model's predicted video category, updating the category wherever needed. The \medvidqa{} dataset contains the collection of  $3,010$ manually created health-related questions and timestamps as visual answers to those questions from trusted video sources, such as accredited medical schools with an established reputation, health institutes, health education, and medical practitioners. We have provided a schematic overview of building the \medvidqa{} and \medvidcl{} datasets in Fig \ref{fig:medvidqa_dataset_overview} and \ref{fig:medvidcl_dataset_overview}, respectively. We benchmarked the datasets by experimenting with multiple algorithms for video classification and video localization.

\section{Methods}
\subsection{MedVidQA Data Creation} \label{medvidqa_methods}
To create the \medvidqa{} dataset, we follow a systematic approach that involves the contributions of medical informatics experts at multiple stages. The detailed steps to build the dataset are as follows:
\begin{enumerate}
    \item \textbf{Extraction of Medical and Health-related Tasks from WikiHow:} 
    With an aim to obtain medical instructional videos that describe how to perform certain health-related activities, we first start by compiling an extensive list of health-related activities using WikiHow\footnote{\url{https://www.wikihow.com/Main-Page}} – an online resource that contains $235,877$ articles on how to do a certain task for a variety of domains ranging from computer and electronics to philosophy and religion, structured in a hierarchy. We start with extracting the medical and health-related tasks from the WikiHow. We collected a total of $6,761$ how-to tasks from the WikiHow `{Heath}' category. 
    \item \textbf{Identification of Relevant Health-related Tasks:} 
In the second step of the dataset creation, we filter the compiled collection extracted from WikiHow. A medical informatics expert reviews each WikiHow task and marks them as relevant or non-relevant from a medical instructional perspective. We keep only those tasks for which the textual answer will be hard to understand and act upon without visual aid such as ``\textit{how to relieve multiple myeloma pain}'' or ``\textit{how to apply cold therapy}''. This procedure yields $1,677$ medical and health-related instructional tasks. 
    \item \textbf{Searching YouTube for The Relevant Videos:} 
    To acquire the relevant videos, we use the task name as a query to search YouTube via its Data API\footnote{\url{https://developers.google.com/youtube/v3}}. In order to collect only the most relevant videos, we only keep the top-4 videos returned by YouTube. We deduplicate videos based on YouTube IDs because some videos may appear in multiple health-related instructional tasks. However, if a video was uploaded multiple times or edited and re-uploaded, the dataset may still contain duplicates.
    \item \textbf{Expert Annotation for Medical Instructional Videos:} 
Medical informatics experts further categorize the relevant medical instructional videos retrieved from YouTube searches. We perform this important step in our dataset creation because \textbf{(1)} videos retrieved by the YouTube search may not be instructional for particular medial queries, and \textbf{(2)} to ensure the reliability of the medical videos. To identify the medical instructional videos in a pool of YouTube videos, we define medical instructional videos as follows:  a medical instructional video should clearly demonstrate a medical procedure providing enough details to reproduce the procedure and achieve the desired results. The accompanying narrative should be to the point and should clearly describe the steps in the visual content. 
If a valid medical query is aligned with an instructional medical video, it should explain/answer the medical query with a demonstration, be a tutorial/educational video where someone (e.g., a \textit{doctor} or a \textit{medical professional}) demonstrates a procedure related to the medical query, or be a how-to video about the\textit{ medical query}. Medical instructional videos may target different levels of expertise, ranging from good Samaritans providing first aid to medical students learning a procedure or experienced clinicians interested in continuing medical education. For this study, we focus on the instructional medical videos that do not require medical education, i.e., the instructions should be at a level that is understandable and can be performed by a layperson. For example, if a nurse shows how to bandage a wound in an emergency, the video is instructional for a layperson. Conversely, if a doctor explains how to perform a specific surgical procedure, the video is instructional for professionals but not for the general public.
\item \textbf{Formulating Instructional Question and Visual Answer from Videos:}
With the aim of formulating medical and health-related questions and localizing their visual answer in the videos, we start with the medical instructional videos annotated in the previous step of the dataset creation. A question is called instructional if the answer requires a step-by-step demonstration and description of the actions to be taken to achieve the goals. For many medical questions, the answer to the question is better shown than described in words, and the answer will be hard to understand and act upon without visual aid, e.g.,
\textit{``how to perform a  physical examination for breast abnormalities?''} Three medical informatics experts were asked to formulate the medical and health-related instructional questions by watching the given video and localizing the visual answer to those instructional questions by providing their timestamps in the video. We asked the annotators to create questions for which \textbf{(1)} answers are shown or the explanation is illustrated in the video, 
\textbf{(2)} the given video is necessary to answer the question, and 
\textbf{(3)} the answer cannot be given as text or spoken information without visual aid.

\end{enumerate}

\subsection{MedVidCL Data Creation} \label{medvidcl_methods}
A video question answering system that can provide the visual answers to medical or health-related instructional questions must have the capability to distinguish between medical instructional and non-instructional videos related to the user's questions. Towards building systems to perform this task efficiently and effectively, we created the \medvidcl{} dataset, which can be used to train a system that can distinguish amongst the medical instructional, medical non-instructional, and non-medical videos. The details of the approach to build \medvidcl{} dataset are described as follows:
\begin{enumerate}[label=(\Alph*)]
 \item \textbf{Collecting Medical and Non-medical Videos}:  With an aim to reduce the annotation efforts, we follow a two-step process to build the \medvidcl{} dataset. In the first step, we seek a high confidence opinion on the video category from a pre-trained video classifier. In the second step, medical informatics experts validate the video category predicted by the video classifier. In order to train the video classifier, we begin with collecting medical and non-medical videos. We utilized the human-annotated $1,016$ medical instructional and $2,685$ medical non-instructional videos from \medvidqa{} dataset. To collect non-medical videos, we sampled $1,157$ videos of  non-medical categories (Food and Entertaining, Cars \& Other Vehicles, Hobbies and Crafts, Home and Garden	Tools, etc.) from the HowTo100M \cite{miech2019howto100m} dataset, which is a large-scale YouTube video dataset with an emphasis on instructional videos. We perform a stratified split on this collection and used $80\%$ videos for training, $10\%$ for validation, and $10\%$ for testing.
 \item \textbf{Building Video Classifier:} We focus on only coarse-grained (\textit{medical instructional}, \textit{medical non-instructional} and \textit{non-medical}) categorization of the videos as opposed to the fine-grained (\textit{walking}, \textit{running}, \textit{playing}, \textit{standing}, etc.) video classification \cite{karpathy2014large} where the micro-level human activity recognition is the key to correctly categorizing the video. Therefore, we hypothesized it is possible to predict the coarse-grained category from the natural language subtitles of the video. Towards this, we propose an ensemble classifier that aggregates the predictions of deep learning and statistical classifiers.  
 We used the support vector machine (SVM) \cite{cortes1995support} as the statistical classifier in our ensemble learning setup, and we chose the pretrained BigBird \cite{zaheer2020big} model as our deep learning classifier as BigBird is capable of accommodating longer sequences that are ubiquitous in video subtitles. We utilized the Hugging Face's implementation\footnote{\url{https://huggingface.co/google/bigbird-roberta-large}} of the BigBird model. After extracting the English video subtitles using the Pytube library\footnote{\url{https://pypi.org/project/pytube/}}, we fine-tuned four different pretrained BigBird models, each with $1024$ as the maximum token length on our training dataset. We also used early stopping to prevent overfitting the model. Since our training dataset has a skewed distribution of the classes, we consider penalizing the model in the training phase for the misclassification made for the minority class by setting higher class weight and at the same time reducing the weight for the majority class. For the class $c \in C$, we set the weight $w_c=\frac{N}{|C|\times N_c}$, where $C$ is the set of all classes in the dataset and $N$ is the total number of samples in the dataset. $N_c$ is the number of samples associated with class $c$ in the dataset.
 We follow the Population Based Training (PBT) \cite{jaderberg2017population} mechanism to jointly optimize a population of models and their hyperparameters to maximize performance. PBT inherits the idea of \textit{exploitation} and  \textit{exploration} from genetic algorithms. In PBT, each member of the population \textit{exploits} -- taking into account the performance of the whole population, a member can decide whether it should abandon the current solution and focus on a more promising one -- and \textit{explores} -- considering the current solution and hyperparameters, it proposes new ones to better explore the solution space. Following this, we fine-tune the BigBird with the PBT strategy (population size=$5$) and consider the top-2 performing members of the population as final models. We used two different PBT strategies to train the BigBird model. In one strategy, we penalize the model and call their top-2 members of the population as \texttt{BigBird$_{\text{w/ weight}}^{1}$} and \texttt{BigBird$_{\text{w/ weight}}^{2}$}. In another strategy, we train the BigBird models without penalizing them and call them \texttt{BigBird$_{\text{w/o weight}}^{1}$} and \texttt{BigBird$_{\text{w/o weight}}^{2}$}.
We adopted the Linear SVC implementation\footnote{\url{https://scikit-learn.org/stable/modules/generated/sklearn.svm.LinearSVC.html}} with the default hyper-parameters settings to train the SVM classifier on our training dataset. We used majority voting from predictions of all five different (4 BigBird + SVM) models in our ensemble learning setting. We break the ties with predictions from the best-performing classifier. The detailed video classification results are shown in Table \ref{tab:ensemble-classifier-results}.
\item \textbf{Identification of Relevant Videos} : We sampled a subset of videos from the large-scale HowTo100M and YouTube8M \cite{abu2016youtube} datasets, and we only choose medical and health-related videos from a set of predefined categories marked as appropriate by medical informatics experts. This process yields a collection of $66,312$ videos from HowTo100M and YouTube8M datasets.
\item \textbf{Predicting Relevant Video Categories from the Video Classifier}: We utilized the ensemble classifier to predict the category of the relevant videos. The ensemble classifier predicted $13,659$ medical instructional videos, $5,611$ medical non-instructional videos, and $47,042$ non-medical videos from the collection of $66,312$ relevant videos.

\item \textbf{Sampling High-Quality Videos and their Manual Assessment}: In order to create a high-quality dataset, we only chose the videos for which the classifier confidence was high for a specific video category and filtered out the videos for which the ensemble classifier confidence was low. In the first step, we filtered out all the videos from the predicted medical-instructional category for which the classifier confidence was below $0.8$. A similar strategy was used for medical non-instructional (confidence score below $0.9$) and non-medical (confidence score below $0.99$). The second and final step involved the manual assessment of the classifier predicted videos.
\end{enumerate}

\begin{figure}[h]
\centering
\begin{minipage}{.45\textwidth}
  \centering
  \includegraphics[height=5cm]{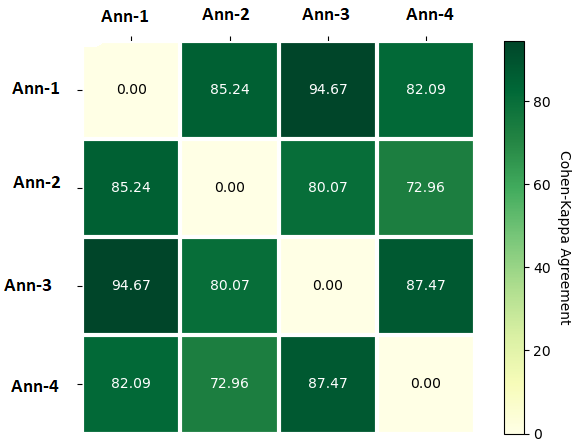}
  \captionof{figure}{Heatmap for the Cohen's kappa based inter-annotator agreements.}
  \label{fig:medvidqa-video-selection-cohen-kappa}
\end{minipage}%
\begin{minipage}{.45\textwidth}
  \centering
  \includegraphics[height=6cm]{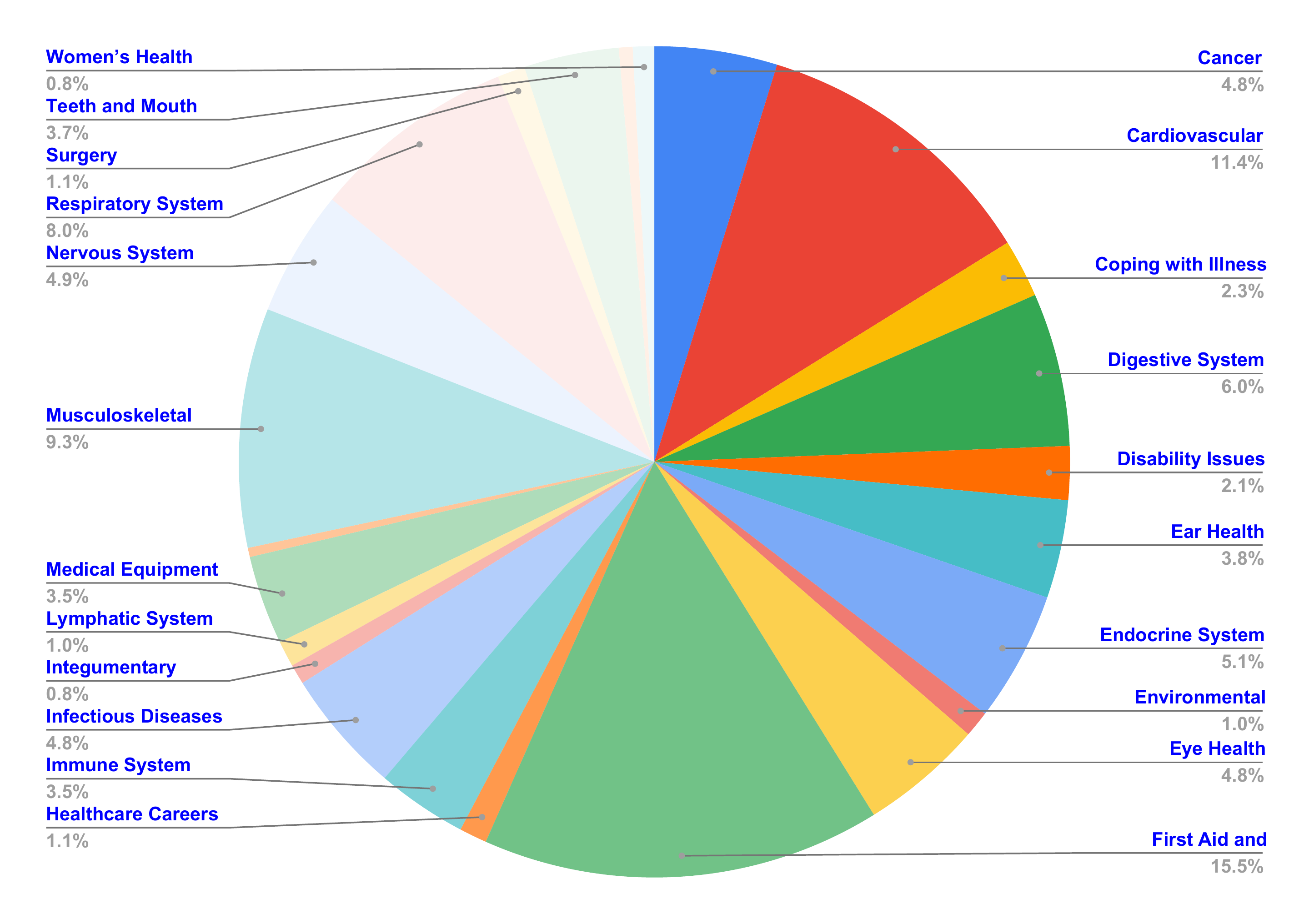}
  \captionof{figure}{Distribution of the instructional tasks category selected from the WikiHow.}
  \label{fig:wikihow-cat}
\end{minipage}%
\end{figure}

\begin{figure}[h]
\begin{minipage}{.55\textwidth}
  \centering

  \resizebox{\textwidth}{!}{%
\begin{tabular}{l|ccc}
\hline
\textbf{Model} & \textbf{Precision} & \textbf{Recall} & \textbf{F1-score} \\
\hline
      \texttt{BigBird$_{\text{w/o weight}}^{1}$}  & $93.27$          & $93.16$       & $93.17$         \\
     \texttt{BigBird$_{\text{w/o weight}}^{2}$} & $94.65$          & $92.66$       & $93.60$         \\
     \texttt{BigBird$_{\text{w/ weight}}^{2}$} & $93.53$          & $91.79$       & $92.62$         \\
     \texttt{BigBird$_{\text{w/ weight}}^{2}$} & $92.61$          & $92.12$       & $92.36$         \\
      SVM & $94.39$      & $91.13$       & $92.57$         \\ \hline
  Ensemble     & $\textbf{95.07}$          & $\textbf{93.65}$       & $\textbf{94.33}$      \\  \hline
\end{tabular}%
}
  \captionof{table}{Performance comparison (on test dataset) of the different video classifiers used in creating \medvidcl{} dataset. All reported results demonstrate the macro average performance.}
  \label{tab:ensemble-classifier-results}

\end{minipage}
\centering
\quad
\begin{minipage}{.4\textwidth}
  \centering
\resizebox{\textwidth}{!}{%
\begin{tabular}{c|ccc|c}
\hline
\textbf{Video Category} & \textbf{Train} & \textbf{Validation} & \textbf{Test} & \textbf{Total}  \\
\hline
      Medical Instructional  & $789$          & $100$       & $600$   &  $1,489$     \\
      \begin{tabular}[c]{@{}c@{}}Medical  Non-instructional\end{tabular} & $2,394$          & $100$       & $500$ & $2,994$         \\
     Non Medical      & $1,034$   & $100$    & $500$   & $1,634$      \\  \hline
      Total      & $4,217$   & $300$    & $1,600$  & $6,117$  \\ \hline
\end{tabular}%
}
  \captionof{table}{Detailed class-wise statistics of the \medvidcl{} dataset.}
  \label{tab:medvidcl_data_stats}
\end{minipage}
\end{figure}

\begin{figure}[t]
\centering
\begin{minipage}{.45\textwidth}
  \centering
  \includegraphics[height=5cm]{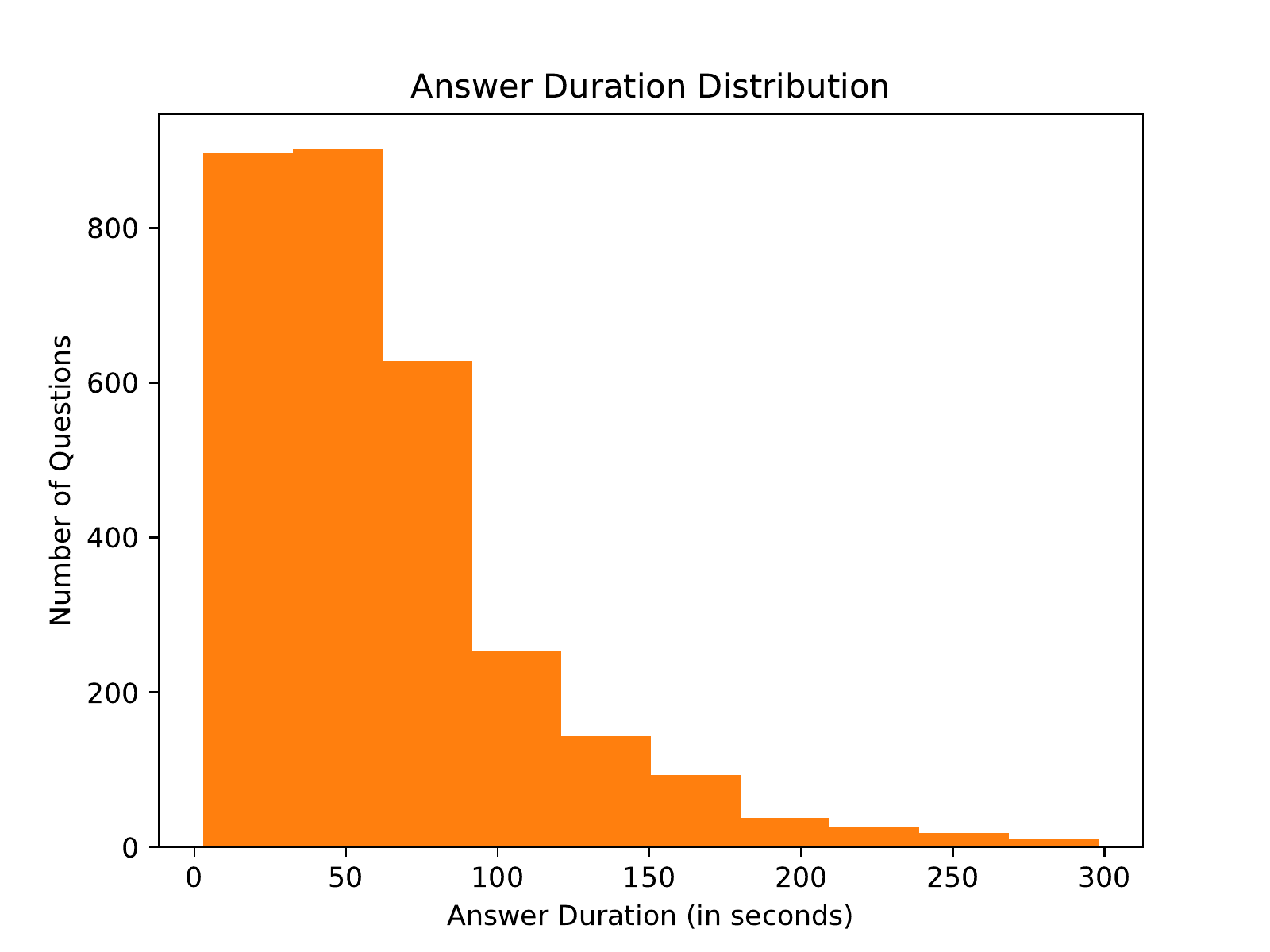}
  \captionof{figure}{Answer duration distribution in \medvidqa{} dataset.}
  \label{fig:medvidqa-answer-distribution}
\end{minipage}%
\quad
\begin{minipage}{.45\textwidth}
  \centering
  \includegraphics[scale=0.4]{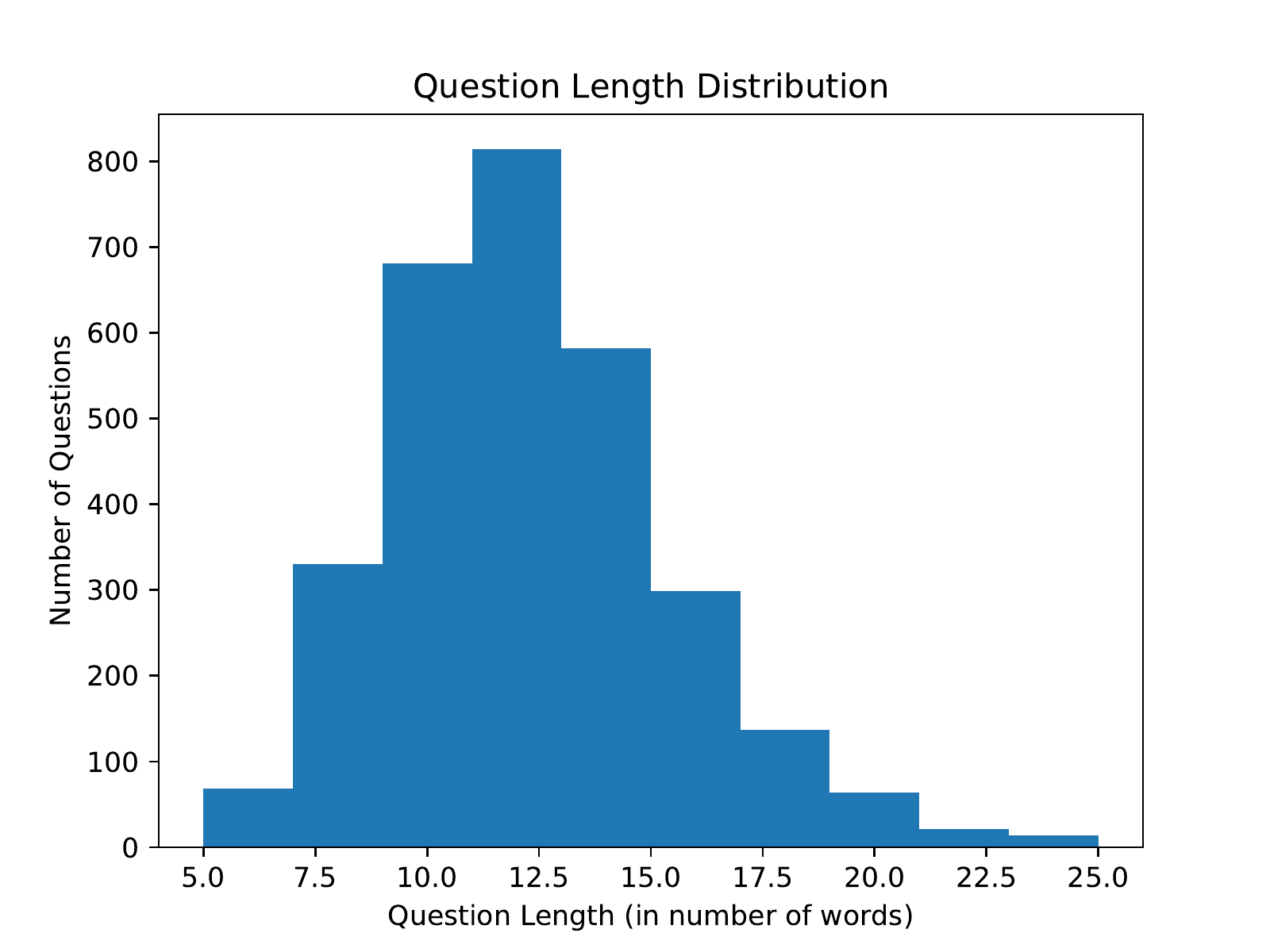}
  \captionof{figure}{Question length distribution in \medvidqa{} dataset.}
  \label{fig:medvidqa-question-distribution}
\end{minipage}%

\end{figure}

\section{MedVidQA Analysis and Validation}
In the first step of the \medvidqa{} dataset creation, we aim to identify and use only trustworthy videos and YouTube channels. A video is categorized as a reliable video if it belongs to a YouTube channel from any of the following sources: \textbf{(a)} accredited medical schools with established reputations, \textbf{(b)} health institutes, \textbf{(c)} health education, \textbf{(d)} hospitals, \textbf{(e)} medical professionals or experts discussing a particular health-related topic, \textbf{(f)} or medical professional appearances and discussions on news channels. 
We have annotated a total of $6,052$ YouTube Videos and categorized $1,016$ as medical instructional, $2,685$ as medical non-instructional, $2,076$ as videos from unreliable video/channel, $140$ as non-medical videos, and $132$ as videos that can not be included in the dataset for other reason. A total of $4$ medical informatics expert annotated these videos. To measure the agreements, we sampled $100$ videos from the whole collection and asked all the annotators to categorize them into either medical instructional or medical non-instructional categories. We computed the pair-wise inter-annotator agreement (Fig. \ref{fig:medvidqa-video-selection-cohen-kappa}) amongst them using the Cohen's kappa coefficient\cite{cohen1960coefficient}, and we found strong agreements (average pair-wise kappa coefficients of $83.75$) amongst them. 

In the second step of \medvidqa{} creation, we focus on creating medical or health-related instructional questions. 
A total of three medical informatics experts formulated these questions and visual answers. This process yielded a total of $3010$ pairs of medical questions and their visual answers from $899$ medical instructional videos totaling $95.71$ hours. We split the videos into training ($800$), validation ($49$), and testing ($50$) sets. 
We have provided the detail statistics in Table \ref{tab:medviqa_data_stats}, Fig. \ref{fig:medvidqa-answer-distribution}, and Fig. \ref{fig:medvidqa-question-distribution}.  To validate the dataset, we sampled $50$ videos and their question-answers annotated by one annotator and asked another annotator to formulate question-answers pairs from the same videos. We {first} manually assessed whether both the annotators had formulated semantically similar questions from the given video, and we then computed the absolute differences between answer timestamps for semantically similar questions formulated by both the annotators. With the first assessment, we measured the number of instances where both the annotators have the same agreement on formulating semantically similar questions from the videos. The second assessment validates their agreements on proving the precise and valid answer timestamps from the videos. We found that both the annotators formulated $93$ and $96$ questions, and $84$ out of them were semantically similar. We computed the average absolute difference (AAD) of start and end timestamps of the visual answers. The AAD values for start and end timestamps are $2.53$ and $3.37$ seconds, respectively. Lower AAD values signify that both annotators consider almost the same answer timestamps whenever they create a semantically similar question. These assessments validate the quality of the \medvidqa{} dataset. 

\section{MedVidCL Analysis and Validation}
To build the \medvidcl{} dataset, we chose human-annotated (`\textit{Medical Instructional}', `\textit{Medical Non-instructional}' and `\textit{Non-medical}') videos from the \medvidqa{} dataset. We considered this set as the training set for the \medvidcl{} dataset. To build a validation and test set, we sampled high confidence videos predicted by a video classifier. To further validate the dataset, we asked the medical informatics expert to review the video category predicted by the model. The expert was asked to \textit{correct} the video category if the video classifier mislabels it and mark the videos as \textit{non-relevant} if there is no conclusive evidence in the videos to label them into any of the video classification categories. The final \medvidcl{} dataset contains $6,117$ videos amongst which $1,489$ are medical instructional, $2,994$ are medical non-instructional and $1,634$ are non-medical. We further removed those videos which have a duration longer than $20$ minutes. We have provided the dataset's detail statistics in Table \ref{tab:medvidcl_data_stats}.

\section{MedVidQA Benchmarking}

We benchmarked the \medvidqa{} dataset by performing a series of experiments using state-of-the-art natural language video localization approaches. We adopt the proposed architecture   \cite{zhang2020vslnet}, which treats localization of the frames in a video as a span prediction task similar to answer span prediction \cite{wang-etal-2017-gated,Seo2017Bidirectional} in text-based question answering.  For a given input question and untrimmed video, we first extracted frames ($16$ frames per second) and obtained the RGB visual features $V=\{v_1, v_2, \ldots, v_n\} \in {R}^{n \times d_v}$ using the 3D ConvNet which was pre-trained on the Kinetics dataset \cite{Carreira2017QuoVA}. We also extracted the word representation of the question  $\{w_1, w_2, \ldots, w_m\} \in {R}^{m \times d_w} $ using Glove embeddings \cite{pennington2014glove}. As done before  \cite{zhang2020vslnet}, we utilized the character embedding $\{c_1, c_2, \ldots, c_m\} \in {R}^{m \times d_c} $ obtained from a convolutional neural network \cite{kim-2014-convolutional} to enrich the word representation and obtained the final question representation as $Q=\{w_1 \oplus c_1, w_2 \oplus c_2, \ldots, w_m \oplus c_m\} \in {R}^{m \times (d_w + d_c)}$. We encoded the question and video features using the feature encoder, which consists of four convolution layers, followed by a multi-head attention layer \cite{vaswani2017attention}. We call the video and question representation obtained from the feature encoder as $\hat{V} \in R^{n \times d}$ and $\hat{Q} \in R^{m \times d}$, respectively. We use the attention flow mechanism \cite{Seo2017BidirectionalAF} to capture the cross-modal interactions between video and question features. 

The attention flow mechanism provides the question-aware video feature representation $\Tilde{V} \in R^{n \times d}$. The answers are located using the span predictor as discussed before \cite{zhang2020vslnet}. Particularly, it uses two unidirectional LSTMs -- the first to predict the start timestamp and another to predict the end timestamp of the answer. The first LSTM, labeled $\text{LSTM}_{s}$, takes the $t^{\text{th}}$ feature $\Tilde{V}_t \in R^{d}$ from $\Tilde{V}$ to compute the hidden state $h_t^{s} = \text{LSTM}_{s}(\Tilde{V}_t, h_{t-1}^{s}) \in R^{d}$. Similarly the second LSTM, labeled $\text{LSTM}_{e}$, computes the hidden state $h_t^{e} = \text{LSTM}_{e}(h_t^{s}, h_{t-1}^{e}) \in R^{d}$. Thereafter, scores for the answer start position ($SC_t^{s} = U \times (h_t^{s} \oplus \Tilde{V}_t) + b$) and end position ($SC_t^{e} = W \times (h_t^{e} \oplus \Tilde{V}_t) + c$) are computed. Here $U \in R^{2d}$ ($W  \in R^{2d}$) and $b  \in R^{2d}$ ($c  \in R^{2d}$) are the weight matrices and biases, respectively. Finally, the probability distributions of the start and end positions are computed by $P_s = softmax(SC^{s}) \in R^{n}$ and $P_e = softmax(SC^{e}) \in R^{n}$.

The network is trained by minimizing the sum of the negative log probabilities of the true start and end answer position by the predicted distributions $P_s$ and $P_e$ averaged over all samples in the batch. The network trained using the span prediction loss is called video span localization (\vslbase{}). We also experiment with the Query-Guided Highlighting (QGH) technique introduced prior \cite{zhang2020vslnet} and call this new network \vslqgh{}. 

With the QGH technique, the target answer span is considered as the foreground and the rest of the video as the background. It extends the span of the foreground to cover its preceding and following video frames. The extension is controlled by the extension ratio $\alpha$, a hyperparameter. An extended answer span aims to cover additional contexts and help the network focus on subtle differences between video frames. We use the $300$ dimensional Glove embeddings and $50$ dimensional character embeddings to obtain the word representation in both the \vslbase{} and \vslqgh{} networks. We also use $1024$ dimensional video features throughout the experiments and hidden state dimensions of $128$ in both the LSTM and Transformer-based encoder. The \vslbase{} and \vslqgh{} networks are trained using AdamW optimizer \cite{loshchilov2018decoupled} for $30$ epochs with an initial learning rate of $0.0001$. The best-performing models are chosen based on the performance (IoU=0.7) on the validation dataset.

\paragraph{Benchmarking Metrics:}  We have evaluated the results using \textbf{(a)} Intersection over Union (IoU) that measures the proportion of overlap between the system predicted answer span and ground truth answer span, and \textbf{(b)} mIoU which is the average IoU over all testing samples. Following a prior protocol \cite{yuan2019find}, we have used ``R@n, IoU = $\mu$'', which denotes the percentage of questions for which, out of the top-$n$ retrieved temporal segments, at least one predicted temporal segment intersects the ground truth temporal segment for longer than $\mu$. In our experiment, we only retrieved one temporal segment; therefore, we have $n=1$. Following previous studies \cite{yuan2019find,zhang2020vslnet}, we have reported $\mu \in \{0.3, 0.5, 0.7\}$ to evaluate the performance of the \vslbase{} and \vslqgh{} models.

\paragraph{Benchmarking Results and Discussion:}

We have performed extensive experiments (\textit{c.f.} Table \ref{tab:medvidqa-results}) with \vslbase{} and \vslqgh{} models to evaluate the \medvidqa{} dataset. We start with the Random Mode approach, which randomly predicts the answer span based on the mode value of the visual answer lengths observed in the validation dataset. We also guess the answer span randomly and call the approach Random Guess. We have reported the results of random prediction on the \medvidqa{} test dataset in Table \ref{tab:medvidqa-results}.

With the \vslbase{} model, we ran multiple experiments by varying the frame length from $400$ to $1400$ to assess its effect on the evaluation metrics. We observe that the \vslbase{} model performs better (except IoU=0.3) with a frame position length of $800$. For IoU=0.3, an FPL (Frame Position Length) value of $1400$ seems to outperform over the other variants of the \vslbase{} model. With an optimal frame length of $800$, we perform our next set of experiments with the \vslqgh{} model.

\begin{figure}[t]
\centering
\begin{minipage}{\textwidth}
\centering
\begin{minipage}{.40\textwidth}
  \centering
  \includegraphics[scale=0.4]{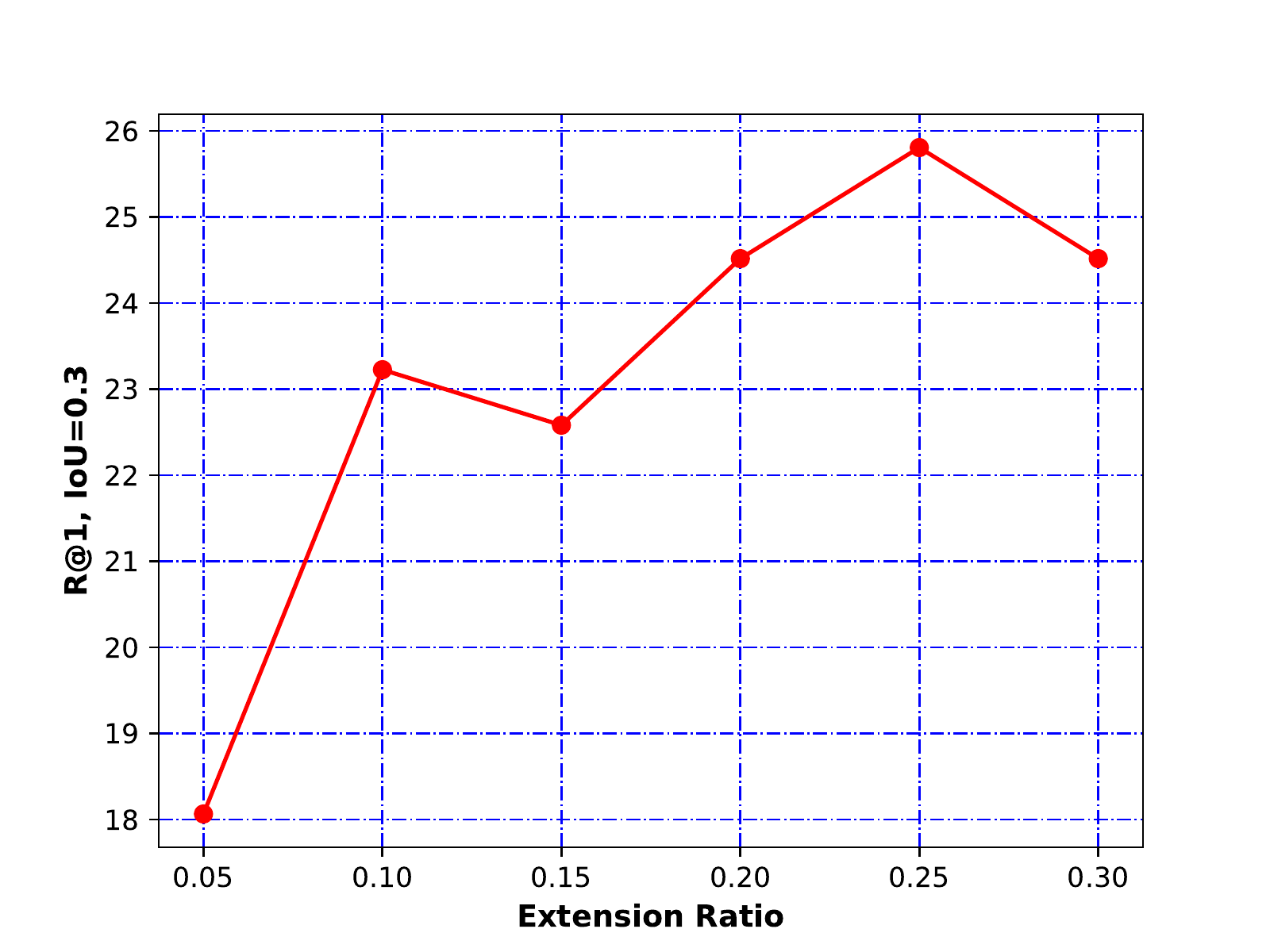}
\end{minipage}%
\begin{minipage}{.40\textwidth}
  \centering
  \includegraphics[scale=0.4]{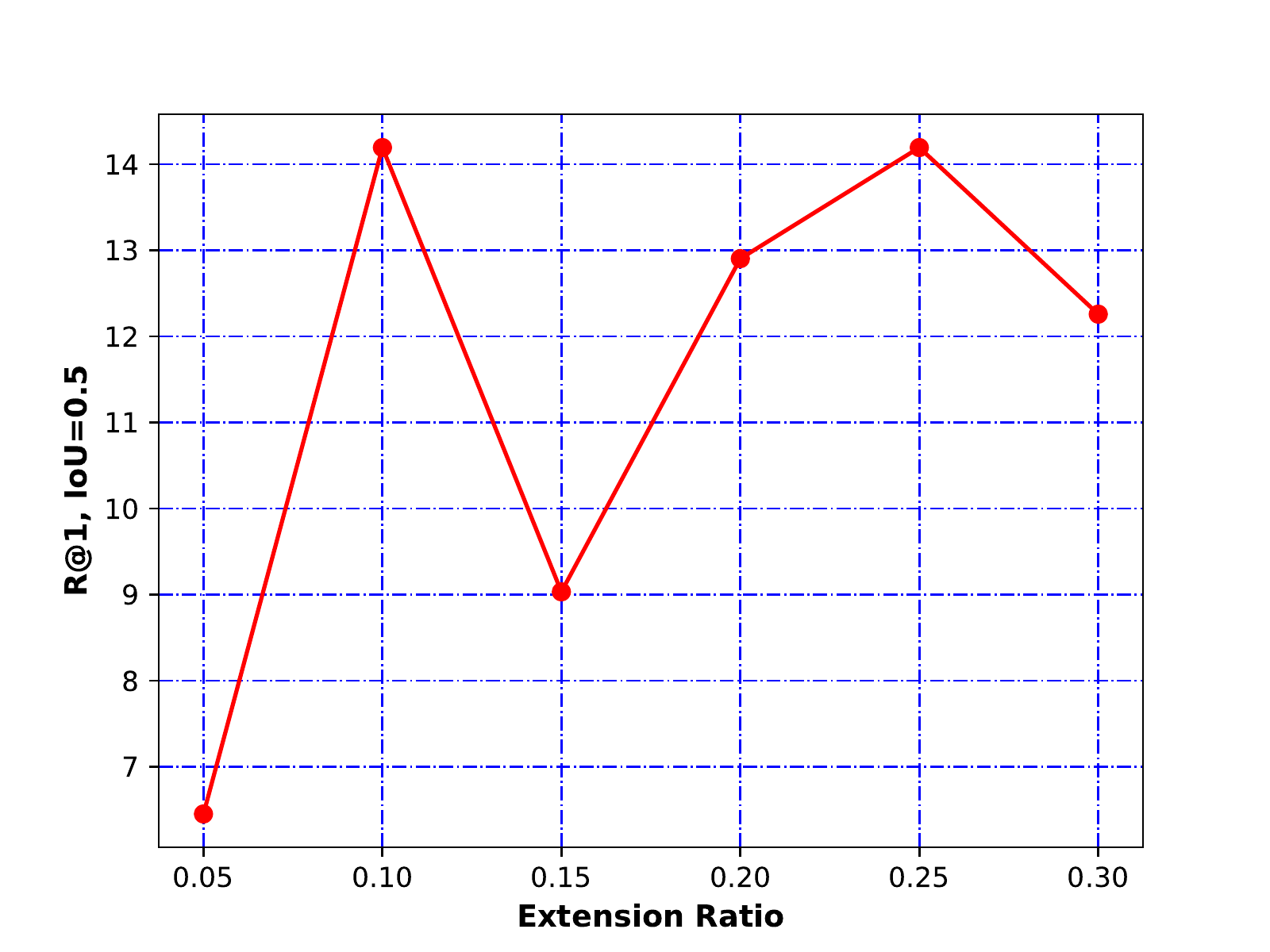}
\end{minipage}%
\end{minipage}%
\newline
\begin{minipage}{\textwidth}
\centering
\begin{minipage}{.40\textwidth}
  \centering
  \includegraphics[scale=0.4]{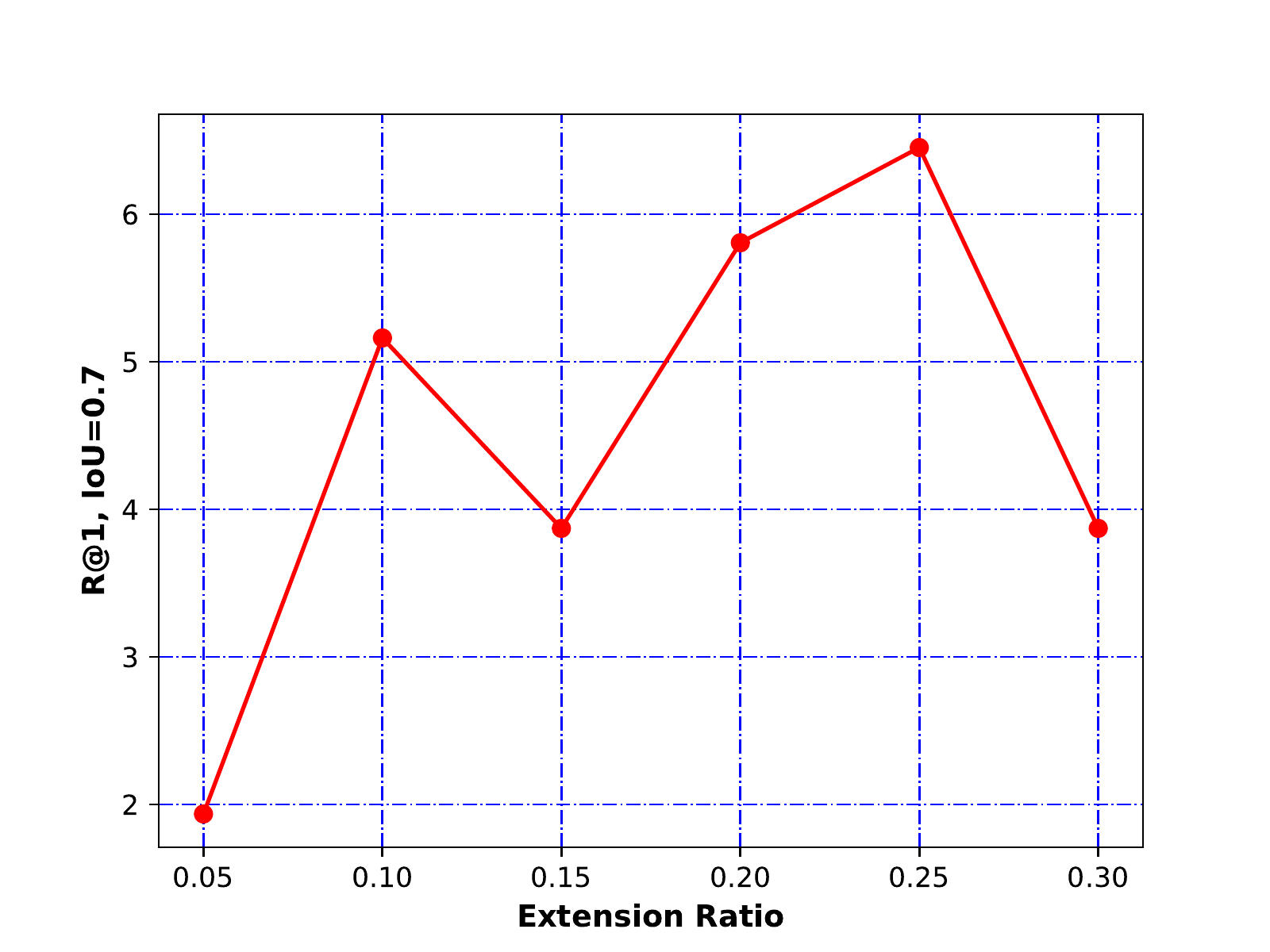}
\end{minipage}%
\begin{minipage}{.40\textwidth}
  \centering
  \includegraphics[scale=0.4]{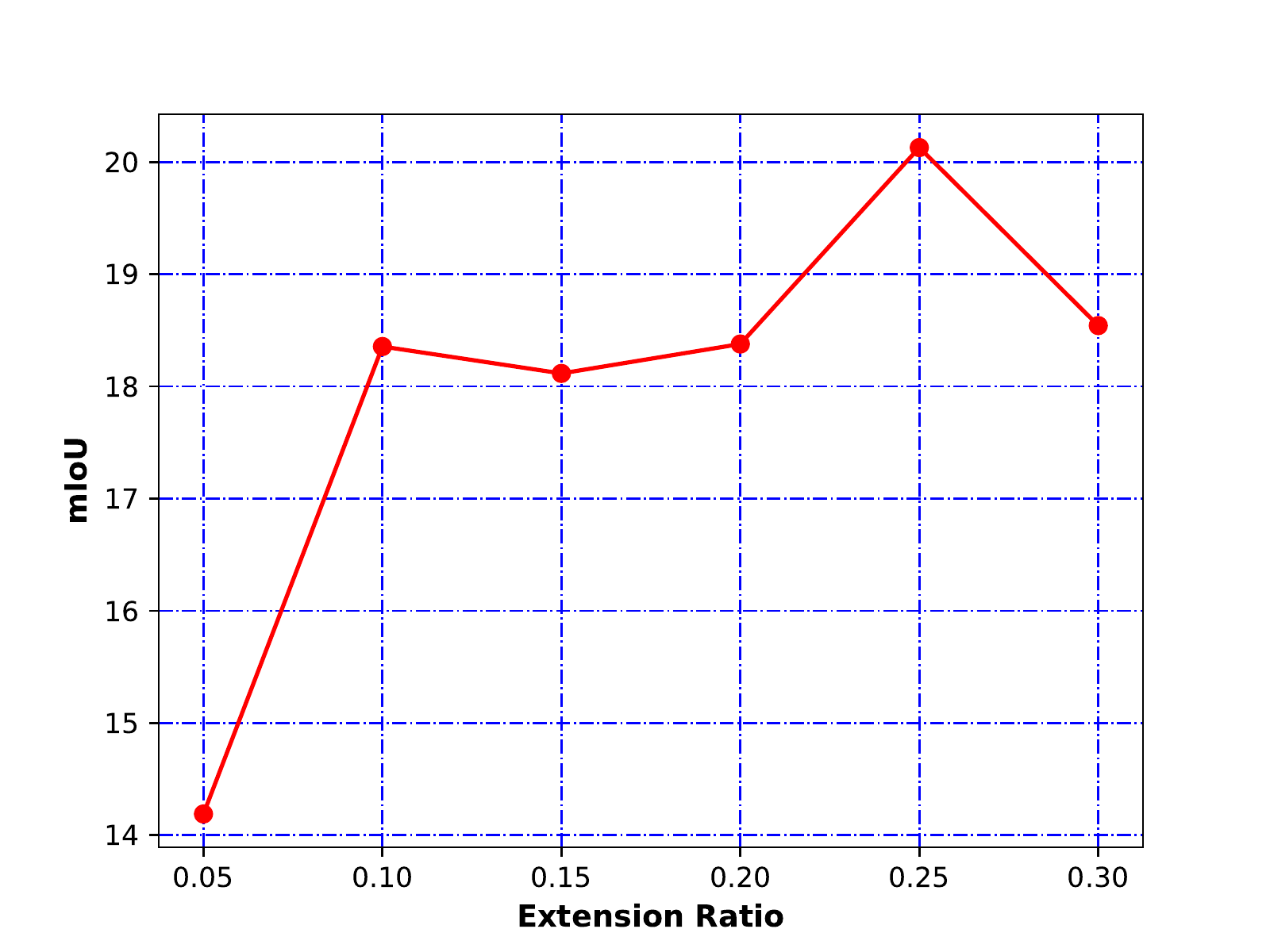}
\end{minipage}%
\end{minipage}%

\caption{Effect of extension ratio ($\alpha$) on the performance of \vslqgh{} model on \medvidqa{} test dataset.}

\label{fig:alpha_analysis}

\end{figure}

The \vslqgh{} models depends on the extension ratio $\alpha$, and the network is trained with join span prediction and visual region (foreground or background) prediction losses. We experiment with the \vslqgh{} model by varying the $\alpha$ from $0.05$ to $0.3$ and reported the results in Fig. \ref{fig:alpha_analysis}. It can be visualized from Fig. \ref{fig:alpha_analysis} that the model outperforms with $\alpha=0.25$. We reported the result for the \vslqgh{} model with its optimal value of $\alpha$ in Table \ref{tab:medvidqa-results}. The \vslqgh{} model obtained the $25.81$ IoU=0.3, $14.20$ IoU=0.5, $6.45$ IoU=0.7, and $20.12$ mIoU. The performance of the \vslqgh{} model in terms of mIoU ($20.12$) is slightly lower ($\downarrow 0.03$) than the \vslbase{} with FPL value of $800$. The results show that the visual answer localization is a challenging task, where the model should have the capability of inter-modal communication to locate the relevant frames in the videos. With multiple useful applications of medical visual answer localization in healthcare and consumer health education, we believe that the \medvidqa{} dataset and benchmarked setup can play a valuable role for further research in this area.

\begin{table}[t]
\centering
\begin{minipage}[]{.75\textwidth}
  \centering
\resizebox{\textwidth}{!}{%
\begin{tabular}{l|c|c|c|c}
\hline
\textbf{Dataset Detail}  &     \textbf{Train}  & \textbf{Validation} & \textbf{Test}    & \textbf{Total}        \\ \hline
Medical instructional videos   &      $800$ &   $49$ & $50$   & $899$        \\
Video duration (hours)         & $86.37$   & $4.54$ & $4.79$              & $95.71$  \\
Mean video duration (seconds)         & $388.68$   & $333.89$ & $345.42$              & $383.29$  \\
Questions and visual answers   & $2,710$ & $145$ & $155$ &$3,010$       \\
Minimum question length                      & $5$   & $6$ & $5$ & $5$       \\
Maximum question length                      & $25$ &  $21$    & $24$    & $25$ \\
Mean question length                   & $11.67$ & $11.76$    & $12$  & $11.81$      \\
Minimum visual answer length (seconds) & $3$ &$10$ &$4$               & $3$     \\
Maximum visual answer length (seconds)  & $298$  &$267$ & $243$             & $298$   \\
Mean visual answer length (seconds)    & $62.29$ & $66.81$    & $56.92$            & $62.23$ \\
Mode visual answer length (seconds)    & $34$ & $36$    & $25$            & $34$ \\

\hline
\end{tabular}%
}
\captionof{table}{Detailed \medvidqa{} dataset statistics for questions, videos, and visual answers. Question length denotes the number of tokens in the questions after performing tokenization with NLTK \cite{bird2004nltk} tokenizer.}
\label{tab:medviqa_data_stats}
\end{minipage}%
\quad
\begin{minipage}[]{.6\textwidth}
  \centering
  \resizebox{\textwidth}{!}{%
\begin{tabular}{l|l|c|c|c|c}
\hline
\multicolumn{2}{c|}{\textbf{Models}}                        & \textbf{IoU=0.3} & \textbf{IoU=0.5} & \textbf{IoU=0.7} & \textbf{mIoU} \\ \hline
 \multicolumn{2}{l|}{\textbf{Random Mode}}                      & 8.38&	1.93&	1.21 &	6.89     \\ 
  \multicolumn{2}{l|}{\textbf{Random Guess}}                      & 7.74    & 3.22            & 0.64            &   5.96\\ \hline
\multirow{5}{*}{\rotatebox[origin=c]{90}{\textbf{\vslbase{}}}}
 & FPL 400                        & 19.35   & 6.45            & 3.22             & 18.08\\    
 & FPL 600                        & 19.35   & 10.96            & 4.51             & 19.20         \\ 
&     FPL 800                         & 21.93   & 12.25            & 5.80             & \textbf{20.15}         \\ 
 &   FPL 1000                           & 21.93   & 7.74             & 3.87             & 18.86         \\ 
 &   FPL 1200                           & 22.58   & 9.67             & 5.16             & 19.97         \\ 
 &  FPL 1400                         & 25.16   & 8.38             & 4.51             & 19.33         \\ \hline
\multicolumn{2}{l|}{\textbf{\vslqgh{}}}            & \textbf{25.81 }  & \textbf{14.20 }           & \textbf{6.45  }           & 20.12         \\ \hline
\end{tabular}%
}
\captionof{table}{Performance comparison of the variants of VSL models on \medvidqa{} dataset. Here \textbf{FPL} refers to the frame position length considered during training the respective models. 
}
\label{tab:medvidqa-results}
\end{minipage}%
\end{table}

\section{MedVidCL Benchmarking}
We benchmarked our \medvidcl{} dataset with multiple monomodal and multimodal approaches. For monomodal approaches, we built several models by considering the language (video subtitle) and vision (video frames) separately for different models. To develop language-based monomodal approaches, we extracted the English subtitles from the videos using the Pytube library\footnote{\url{https://pypi.org/project/pytube/}}. We then trained statistical classifiers such as Linear SVC \cite{fan2008liblinear}, and SVM \cite{cortes1995support} to predict the video category by considering the language features. We have also experimented with pre-trained language models such as BERT-Base-Uncased \cite{devlin2019bert}, RoBERTa-Base \cite{liu2019roberta} and BigBird-Base \cite{zaheer2020big}. For vision-based monomodal approaches, we extracted $20$ frames from each video at a uniform time interval and used these frames as the sequence information to predict the video category. To process these frames, we utilized a 3D ConvNet (I3D), which was pre-trained on the Kinetics dataset \cite{Carreira2017QuoVA}, and the resulting features were passed as input to both LSTM and Transformer networks to predict the video category. We have also experimented with the pre-trained\footnote{\url{https://huggingface.co/google/vit-base-patch16-224}} ViT\cite{dosovitskiy2020vit} model for extracting frames and obtained the frame representation from the ViT feature extractor. Similar to the I3D, we passed the resulting features to LSTM and Transformer networks to predict the video category. We extended our experiments from monomodal to multimodal settings by considering both the language and vision features. Using the language input (video subtitle) and vision input (video frames), we obtained their representations either from LSTM or Transformer networks. We then concatenated the language and vision features together and pass the concatenated features to a feed-forward layer to predict the video category. Similar to the monomodal (vision) experiments, we use both I3D and ViT features to perform the multimodal experiments.

We chose the hyper-parameters values based on the classifier performance (average macro F1-score) on the validation dataset. For the Linear SVC and SVM classifiers, the optimal regularization value $C$ was $1.5$ and $1$, respectively. The SVM model with the \textit{sigmoid} kernel outperformed the other kernels on the validation dataset. We utilized the pre-trained Transformer models from Hugging Face\cite{wolf-etal-2020-transformers} to perform the monomodal (language) experiments. Each pre-trained Transformer model was trained with the AdamW optimizer with the learning rate=$5e-5$ for $10$ epochs and with the batch size of $8$ (except for BigBird, where the batch size was $4$). The LSTM and Transformer hidden states ertr set to $128$ and $1024$, respectively, for all the monomodal (vision) and multimodal experiments. Each monomodal (vision) and multimodal network was trained with the learning rate=$5e-5$ for $30$ epochs with the batch size of $16$. We set the maximum text sequence length for multimodal experiments to $512$.
\paragraph{Benchmarking Metrics: }  The evaluation metric to assess the performance of the systems are \textbf{(a)} F1-score on Medical Instructional Class, and \textbf{(b)} macro average F1-score across all the classes.

\begin{table}[]
    \centering
   \resizebox{0.85\textwidth}{!}{%
\begin{tabular}{cl|c|c|c||c|c|c}
\hline
\multicolumn{2}{c|}{\textbf{Models}}      & \textbf{Precision} & \textbf{Recall} & \textbf{F1-score} & 
 \begin{tabular}[c]{@{}c@{}} \textbf{Precision}\\ \textbf{ (Med-Inst)} \end{tabular}
 & \begin{tabular}[c]{@{}c@{}} \textbf{Recall}\\ \textbf{ (Med-Inst)} \end{tabular} & \begin{tabular}[c]{@{}c@{}} \textbf{F1-score}\\ \textbf{ (Med-Inst)} \end{tabular} \\ \hline
\multicolumn{1}{c|}{\multirow{5}{*}{\textbf{\begin{tabular}[c]{@{}c@{}}Monomodal\\ (Language)\end{tabular}}}} &
  Linear SVC \cite{fan2008liblinear} &
  89.64 &
  89.71 &
  88.41 &
  99.76 &
  70.33 &
  82.50 \\ 
\multicolumn{1}{c|}{} & SVM \cite{cortes1995support}               & 89.54     & 88.73  & 87.42   & \cellcolor{lightgreen}100.0         & 67.00        & 80.24      \\ 
\multicolumn{1}{c|}{} & BERT-Base-Uncased \cite{devlin2019bert} & 92.82     & 93.23  & 92.91   & 95.98         & 87.50        & 91.54      \\ 
\multicolumn{1}{c|}{} & RoBERTa-Base \cite{liu2019roberta}      & 94.58     & 94.98  & 94.67   & 97.99         & 89.33        & 93.46      \\  
\multicolumn{1}{c|}{} & BigBird-Base \cite{zaheer2020big}      & \cellcolor{lightgreen}95.58     & \cellcolor{lightgreen}95.96  & \cellcolor{lightgreen}95.68   & 98.19         & \cellcolor{lightgreen}90.67        & \cellcolor{lightgreen}94.28      \\ \hline \hline
\multicolumn{1}{c|}{\multirow{4}{*}{\textbf{\begin{tabular}[c]{@{}c@{}}Monomodal\\ (Vision)\end{tabular}}}} &
 I3D + LSTM \cite{Carreira2017QuoVA,hochreiter1997long} &
 75.62	& 75.88	 & 75.11	&81.66 &	63.83	& 71.66 \\
\multicolumn{1}{c|}{} & ViT + LSTM \cite{dosovitskiy2020vit,hochreiter1997long}  &
\cellcolor{naplesyellow}82.07	& 81.16 &	80.49	& \cellcolor{naplesyellow}89.62 &	67.67 &	77.11 
\\
\multicolumn{1}{c|}{} & I3D + Transformer \cite{Carreira2017QuoVA,vaswani2017attention}   &    75.18 &	75.41&	74.43&	83.14&	60.83&	70.26\\
\multicolumn{1}{c|}{} & ViT + Transformer \cite{dosovitskiy2020vit,vaswani2017attention}     &  81.76 &	\cellcolor{naplesyellow}82.06&	\cellcolor{naplesyellow}81.26&	89.25&	\cellcolor{naplesyellow}69.17&	\cellcolor{naplesyellow}77.93 \\ \hline \hline
\multicolumn{1}{c|}{\multirow{3}{*}{\textbf{\begin{tabular}[c]{@{}c@{}}Multimodal\\ (Language + Vision)\end{tabular}}}} &
  L + V (I3D) + LSTM &
 75.96	& 76.16 &	75.68	& 79.68 &	66.67	& 72.60
  \\  
\multicolumn{1}{c|}{} &
   L + V (ViT) + LSTM &
 82.57	& 82.16 & 81.40	& 90.22 &	67.67	& 77.33
 \\ 
\multicolumn{1}{c|}{} &
   L + V (I3D) + Transformer &
74.74	& 75.10 &	74.80 &	76.23 &	\cellcolor{periwinkle}69.50 &	72.71
\\ 
  
 \multicolumn{1}{c|}{} &
   L + V (ViT) + Transformer &
\cellcolor{periwinkle}83.65	& \cellcolor{periwinkle}83.12 &	\cellcolor{periwinkle}82.38	& \cellcolor{periwinkle}92.22 &	69.17 &	\cellcolor{periwinkle}79.05
 
  \\ \hline
  
  \hline
\end{tabular}%
}
\captionof{table}{Performance of the monomodal and multimodal approaches on \medvidcl{} test dataset. Here \textbf{L} and \textbf{V} denotes the Language and Vision respectively. Precision, Recall and F1-score denotes \textit{macro} average over all the classes. The best results amongst monomodal (language) approaches is highlighted with \colorbox{lightgreen}{green} shade. Similarly, we shown the best results from monomodal (vision) and multimodal are shown in \colorbox{naplesyellow}{yellow} and \colorbox{periwinkle}{purple} shades.}
\label{tab:medvidcl-results}
\end{table}

\paragraph{Benchmarking Results and Discussion:}

We have provided the detailed results of multiple monomodal and multimodal approaches in Table \ref{tab:medvidcl-results}. Among language-based monomodal approaches, BigBird-Base outperforms other methods and achieved $95.68\%$ overall macro average F1-score and  $94.28\%$ F1-score for medical instructional class. Pre-trained Transformer-based models perform better than their counterpart SVM variants. Since BigBird can accommodate and handle longer sequences effectively, which is plausible in video subtitle, it shows better performance than the other pre-trained language models (BERT and RoBERTa). Among vision-based monomodal approaches, the feature representation learned using ViT ($81.26$ overall F1-score) is more effective than I3D ($74.43$ overall F1-score) with Transformer-based frame sequence processing. 
With multimodal approaches, we observed improvements over the respective monomodal (vision) approaches. We observed the maximum improvement of $1.12\%$ overall F1-score with the multimodal approach (L + V (ViT) + Transformer) compared to the monomodal (ViT + Transformer) approach. Similar trends are also observed for Medical Instructional video classification, where we reported an increment of $1.12\%$ F1-score with the multimodal approach (L + V (ViT) + Transformer) compared to the monomodal (ViT + Transformer) approach. For the visual answer localization to the health-related questions, it is essential to predict the relevant medical instructional videos correctly; therefore, we prioritize the system's performance on medical instructional classes compared to the overall video classes. In this case, the F1-score on medical instructional videos is more important than the overall F1-score. We believe that a sophisticated language-vision fusion mechanism will further improve the performance (overall F1-score and Medical Instructional F1-score) of the multimodal approaches.

\section*{Acknowledgements} 
This work was supported by the intramural research program at the U.S. National Library of Medicine, National Institutes of Health.
The content is solely the responsibility of the authors and does not necessarily represent the official views of the National Institutes of Health.
We would like to also thank Anna Ripple, Asma Ben Abacha, and Laura Ingwer Bachor for their annotations and contributions to the pilot stages of this work. 

\bibliographystyle{unsrt}  
\bibliography{sample}

\end{document}